\documentclass[runningheads]{llncs}

\usepackage{eccv}

\usepackage{eccvabbrv}

\usepackage{graphicx}
\usepackage{booktabs}

\usepackage{dirtytalk}
\usepackage{adjustbox}
\usepackage{multirow}
\usepackage{pifont}
\usepackage{amsmath}
\usepackage{colortbl}
\usepackage{listings}
\usepackage{cite}
\usepackage{array}

\usepackage[accsupp]{axessibility}  

\usepackage{hyperref}

\usepackage{orcidlink}

\begin{document}

\title{SurgicalPart-SAM: Part-to-Whole Collaborative Prompting for Surgical Instrument Segmentation}

\titlerunning{SurgicalPart-SAM for Surgical Instrument Segmentation}

\author{Wenxi Yue\inst{1} \and
Jing Zhang\inst{1}\thanks{Corresponding Author.}\and
Kun Hu\inst{1} \and
Qiuxia Wu\inst{2} \and 
Zongyuan Ge\inst{3} \and 
Yong Xia\inst{4} \and 
Jiebo Luo\inst{5} \and 
Zhiyong Wang\inst{1} }

\authorrunning{Yue et al.}

\institute{School of Computer Science, The University of Sydney \and
School of Software Engineering, South China University of Technology \and 
Department of Data Science \& AI, Monash University \and
School of Computer Science, Northwestern Polytechnical University \and 
Department of Computer Science, University of Rochester \\ \email{\{wenxi.yue, jing.zhang1, kun.hu, zhiyong.wang\}@sydney.edu.au, qxwu@scut.edu.cn, Zongyuan.Ge@monash.edu, yxia@nwpu.edu.cn, jluo@cs.rochester.edu}}

\maketitle

\begin{abstract}

The Segment Anything Model (SAM) exhibits promise in generic object segmentation and offers potential for various applications. Existing methods have applied SAM to surgical instrument segmentation (SIS) by tuning SAM-based frameworks with surgical data. However, they fall short in two crucial aspects: (1) Straightforward model tuning with instrument masks treats each instrument as a single entity, neglecting their complex structures and fine-grained details; and (2) Instrument category-based prompts are not flexible and informative enough to describe instrument structures. To address these problems, in this paper, we investigate text promptable SIS and propose \textbf{S}urgical\textbf{P}art-\textbf{SAM} (SP-SAM), a novel SAM efficient-tuning approach that explicitly integrates instrument structure knowledge with SAM's generic knowledge, guided by expert knowledge on instrument part compositions. Specifically, we achieve this by proposing (1) Collaborative Prompts that describe instrument structures via collaborating category-level and part-level texts; (2) Cross-Modal Prompt Encoder that encodes text prompts jointly with visual embeddings into discriminative part-level representations; and (3) Part-to-Whole Adaptive Fusion and Hierarchical Decoding that adaptively fuse the part-level representations into a whole for accurate instrument segmentation in surgical scenarios. Built upon them, SP-SAM acquires a better capability to comprehend surgical instruments in terms of both overall structure and part-level details. Extensive experiments on both the EndoVis2018 and EndoVis2017 datasets demonstrate SP-SAM's state-of-the-art performance with minimal tunable parameters. The code will be available at \url{https://github.com/wenxi-yue/SurgicalPart-SAM}.

\end{abstract}

\section{Introduction}
\label{sec:intro}

\begin{figure*}[!t]
    \centering
    \includegraphics[width=1\textwidth]{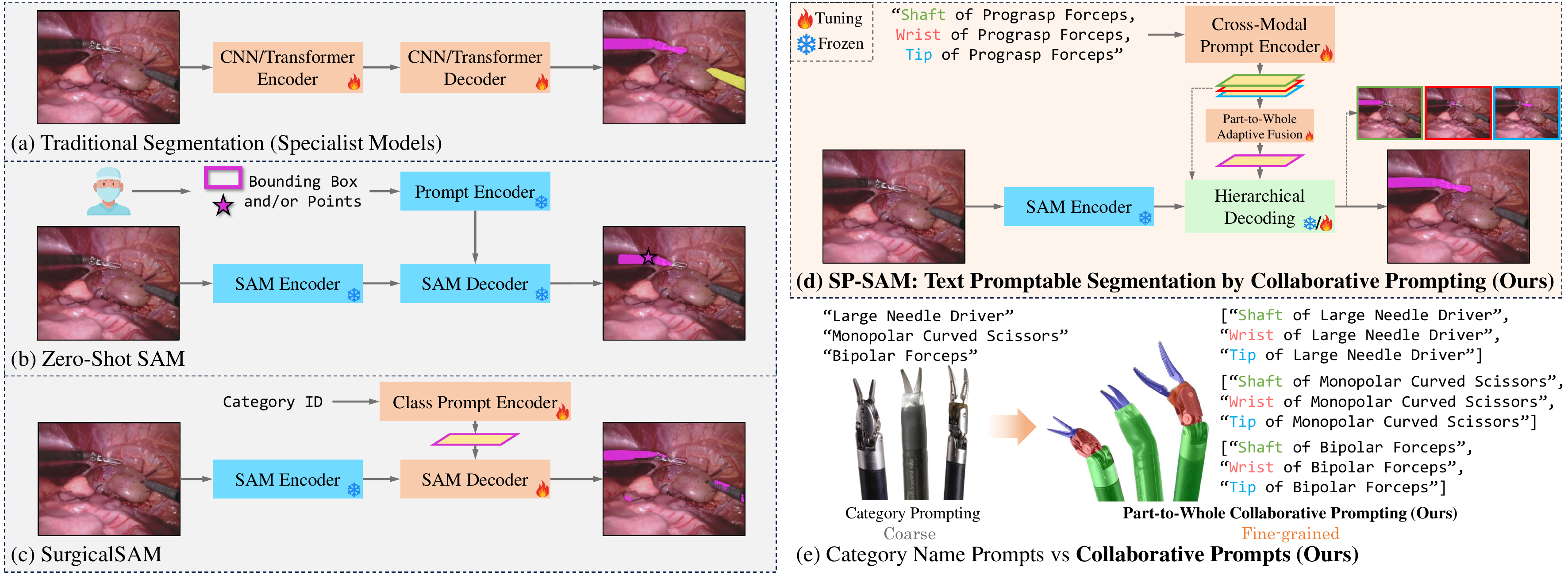}
    \caption{SP-SAM with Collaborative Prompts incorporates the knowledge of surgical instrument structures. Subfigure (e) is partially is excerpted from \cite{endovis2017}.}
   \label{fig:motivation}
 \end{figure*}

Surgical instrument segmentation (SIS) aims to accurately identify and delineate surgical instruments in operative scenes. It plays a foundational role for many downstream applications, such as surgical planning~\cite{planning}, robotic navigation~\cite{navigation}, and skill assessment~\cite{skill}. 
We identify two primary problems with the existing methods for this task (Fig.~\ref{fig:motivation}(a)). 
First, they often develop specialist models~\cite{ternausnet,mftapnet,dual_mf,isinet,trasetr,baby,matis,tpsis} that require training a large number of parameters, leading to high development costs. 
Second, current methods lack the capability of human-computer interaction that is highly desired in surgical practice~\cite{surgery_ar_survey, survey_multimodal_interact}. 

The Segment Anything Model (SAM)~\cite{sam} is a pioneering foundation model for promptable segmentation. It holds great potential for addressing the above problems owing to its rich pre-trained knowledge and interactivity.
However, employing SAM for surgical instrument segmentation in a zero-shot manner (Fig.~\ref{fig:motivation}(b)) poses significant challenges. 
Firstly, zero-shot frameworks of SAM, including detection-based (MaskTrack-RCNN~\cite{masktrackrcnn}/Mask2Former~\cite{m2f} + SAM), tracking-based (TrackAnything~\cite{trackanything}), and reference-based (PerSAM~\cite{persam}) frameworks, have demonstrated inferior generalisation on surgical instruments~\cite{surgicalsam}. This deficiency is mainly due to the insufficient surgical data in SAM pre-training and the notable domain disparity between natural objects and surgical instruments. Specifically, 
compared to generic objects, 
surgical instruments present more intricate structures and fine-grained details, exacerbating the challenge of generalising SAM to this specialised domain.
Secondly, SAM's reliance on point-or-box prompts is impractical in surgical settings, where it is infeasible for surgeons to provide such prompts for every instrument in each frame. 

Initial attempts have been made to address these problems. Yue et al.~\cite{surgicalsam} propose SurgicalSAM (Fig.~\ref{fig:motivation}(c)), an instrument category-prompted SAM framework efficiently tuned with surgical data. Additionally, Wang et al.~\cite{surgical_sam_miccai} propose an efficient-tuning approach for SAM for SIS employing fixed default prompt embeddings. However, these methods suffer from two crucial problems. First, their straightforward tuning approach using whole instrument masks treats each instrument as a single entity and cannot explicitly handle the complex structures and details of instruments. Despite well-established expert knowledge on instrument structure compositions, they fail to incorporate these insights during tuning. Secondly, they depend on instrument category prompts or fixed default prompts, which lack flexibility and intuitiveness for surgeon-computer interaction and fail to provide informative descriptions of instrument structures. Instead, more flexible and informative prompts such as text are preferred.

In this paper, we explore text promptable surgical instrument segmentation and propose a novel framework, \textbf{S}urgical\textbf{P}art-\textbf{SAM} (SP-SAM) (Fig.~\ref{fig:motivation}(d)), to address the above problems. Specifically, we recognise the well-established expert knowledge regarding the compositions of surgical instrument parts, \eg, \textit{Large Needle Driver} is composed of \textit{shaft}, \textit{wrist}, and \textit{tip}, \textit{Monopolar Curved Scissors} is composed of \textit{shaft} and \textit{tip}, etc. In SP-SAM, we aim to harness this expert knowledge to guide the tuning of SAM to improve its capability to comprehend instrument structures and identify subtle details.

To integrate part-level information, we first introduce a new form of text prompt, namely Collaborative Prompts, which utilises a text description set: 
\{[\textit{part name}] of [\textit{instrument category name}]\} for all parts of an instrument category, collaborating category-level and part-level text descriptions. Contrasted with prompting solely with instrument category names, Collaborative Prompts effectively enables the integration of more precise and fine-grained instrument part information (Fig.~\ref{fig:motivation}(e)).
Next, to correlate the Collaborative Prompts with the instrument parts in the image, we introduce a Cross-Modal Prompt Encoder to learn part-level representations via interaction between the Collaborative Prompts and the image embedding. This enables focused learning of fine-grained features for each instrument part, thereby enhancing the segmentation of subtle details. Finally, we propose Part-to-Whole Adaptive Fusion and Hierarchical Decoding to fuse representations of all instrument parts into a whole and decode them into segmentation masks, capturing both the global structure and the compositional components.

Note that, part-to-whole fusion is non-trivial due to two inherent challenges in surgical scenarios: (1) the varying part compositions across instrument categories, and (2) the frequent occlusions of instruments. These challenges necessitate adaptive fusion of different parts for each instrument in the surgical scene. Therefore, within the Part-to-Whole Adaptive Fusion module, we propose Category Part Attention and Image Part Attention. The former adapts category-specific part weightings to accommodate diverse part compositions across categories, while the latter learns adaptive image-specific part weightings to handle occluded or out-of-view parts in the image.
By integrating all components, SP-SAM exhibits a strong capability to adaptively comprehend surgical instrument structures, identify subtle details, and discriminate between fine-grained categories.  
In summary, our contributions are:
\begin{itemize}
        \item We introduce a novel SAM efficient-tuning approach, SurgicalPart-SAM (SP-SAM), for text promptable surgical instrument segmentation. SP-SAM utilises well-established expert knowledge of surgical instrument part compositions to guide SAM tuning, explicitly addressing the structural complexity and subtle details of surgical instruments, thereby enhancing generalisability.

        \item We introduce Collaborative Prompts, Cross-Modal Prompt Encoder, and Part-to-Whole Adaptive Fusion and Hierarchical Decoding, to achieve multi-modal embedding learning at both the part level and the category level. These designs enhance comprehension of instrument structures and details during SAM tuning.

    \item We propose Category Part Attention and Image Part Attention to integrate category-specific and image-specific weights for adaptively fusing instrument part representations. These mechanisms respectively address two critical challenges in surgical scenarios: the varying part compositions across instrument categories and the frequent occlusions of instruments.
    \item We conduct extensive experiments on the challenging EndoVis2018 and EndoVis2017 datasets and show that SP-SAM achieves state-of-the-art performance with only a small number of training parameters. 

\end{itemize}

\section{Related Work}
\label{sec:relatedwork}
\subsection{Surgical Instrument Segmentation}

Most surgical instrument segmentation methods focus on developing specialist models. Early research adopts a semantic segmentation pipeline with the pioneering work TernausNet introducing a U-Net based encoder-decoder model~\cite{ternausnet}. Subsequent developments include feature pyramid attention~\cite{paanet} and flow-based temporal priors~\cite{mftapnet, dual_mf}. An alternative strategy to semantic segmentation is instance segmentation. ISINet adopts Mask-RCNN~\cite{isinet, mask_rcnn} for this task, which is later enhanced by Baby et al.~\cite{baby} with a specialised classification module. In addition, TraSeTR utilises a track-to-segment transformer with tracking cues~\cite{trasetr} and MATIS employs Mask2Former with a temporal consistency module~\cite{matis, m2f}. Recently, Zhou et al.~\cite{tpsis} introduce TP-SIS, a text promptable framework exploiting the pre-trained vision-language model CLIP~\cite{clip}. Despite the variety of specialist models, they all involve fully training a complete set of model parameters, resulting in high development costs.

To enhance model generalisability and reduce training costs, there is a growing interest in adapting pre-trained foundation models for SIS. SurgicalSAM is proposed as a category-prompted SAM framework efficiently tuned with surgical data~\cite{surgicalsam}, while Wang et al.~\cite{surgical_sam_miccai} propose an efficient-tuning method for SAM using fixed default embeddings as prompts. 
However, these approaches rely on less informative prompts and overlook the intricate structures and subtle details of surgical instruments during SAM tuning. In contrast, our SP-SAM employs more informative Collaborative Prompts in text form to explicitly leverage expert knowledge of instrument part compositions to guide the tuning of SAM, enhancing SAM's comprehension of surgical instruments compared to \cite{surgicalsam,surgical_sam_miccai}.

\subsection{Text Promptable Segmentation}
In contrast to traditional segmentation that solely relies on pre-defined class labels, text promptable segmentation uses natural language as prompts that can offer richer contextual information, improved generalisation, and greater flexibility.  Early works are primarily based on Convolutional Neural Networks and Recurrent Neural Networks and propose attention mechanisms for extracting and relating visual and textual features~\cite{tseg_cnn, tseg_rnn, tseg_keyword, tseg_selfatt}. More recent approaches utilise Transformers to perform feature extraction and multi-modal feature fusion~\cite{tseg_coattention, tseg_refSegformer, tseg_restr, tseg_lavt, tseg_querygenerate}. Recently, to leverage the rich knowledge from large-scale pre-training, large vision-language models such as CLIP~\cite{clip} are utilised for this task~\cite{cris, Clipseg}. Zhou et al.~\cite{tpsis} employ CLIP~\cite{clip} and introduce TP-SIS, the first text promptable framework for SIS. However, in TP-SIS~\cite{tpsis}, instrument part masks are used straightforwardly as supervisory signals, neglecting the structural dependencies associated with the parts. Moreover, TP-SIS requires fine-tuning the entire CLIP Image Encoder, resulting in high training costs. In contrast, our SP-SAM explicitly explores category-specific and image-specific part dependencies by incorporating expert knowledge on instrument structures and requires only a very small number of training parameters.

\subsection{Segment Anything Model}
SAM is recognised as the pioneering foundation model for image segmentation. Owing to extensive pre-training on large-scale data, SAM exhibits impressive generalisation capabilities on various downstream tasks~\cite{sam}. However,  its zero-shot performance in medical contexts tends to fall short due to the significant disparity between natural and medical subjects ~\cite{sam_wsi, sam_12datasets, sam_mia, sam_yangxin, sam_ppt_modes, surgicalsam}. Moreover, SAM's reliance on precise per-frame point-or-box prompts for segmentation~\cite{sam_ppt_modes, sam_md} requires extensive manual input, infeasible in many medical scenarios, \eg during surgery. To mitigate the gap between natural and medical domains, some studies have fine-tuned SAM with domain-specific data. However, these methods either have limited interactivity~\cite{sam_ada_cust, sam_ada_udpf, surgical_sam_miccai}, require labour-intensive per-frame points or bounding boxes for prompting ~\cite{medsam, msa}, or rely on inflexible category IDs~\cite{surgicalsam}. 
In contrast to these approaches, in SP-SAM we propose Collaborative Prompts that integrate category-level and part-level texts. This method offers a more intuitive and flexible approach for surgeon-computer interaction, enables informative descriptions of instrument structures, and introduces additional cues to SAM from the language modality.

\section{Method}
\label{sec:method}

\begin{figure*}[!t]
    \centering
      \includegraphics[width=1\textwidth]{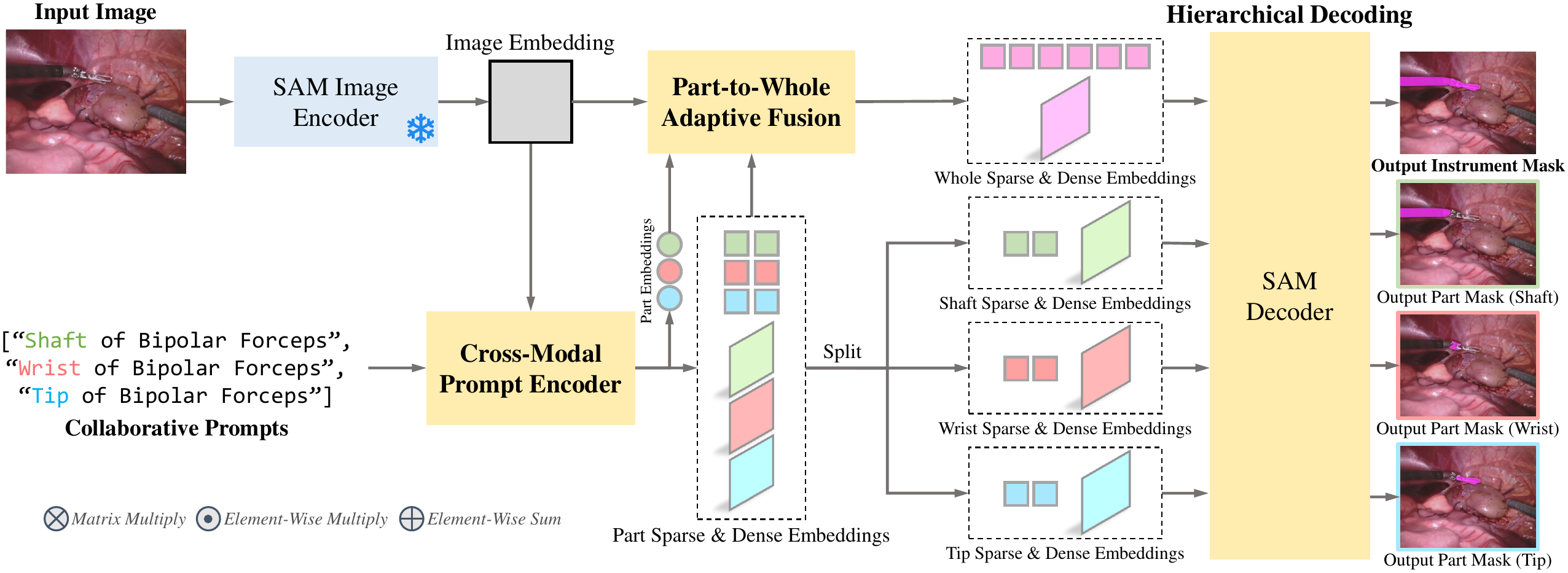}
    \caption{\textbf{Overview of SP-SAM}. SP-SAM consists of four main components: a SAM Image Encoder, a Cross-Modal Prompt Encoder, a Part-to-Whole Adaptive Fusion module, and a SAM Decoder. The SAM Image Encoder, CLIP Text Encoder (within the Cross-Modal Prompt Encoder), and output MLPs in SAM Decoder are frozen and the remaining weights are tuned. 
    }
   \label{fig:overall}
 \end{figure*}

In this paper, we address the task of text promptable surgical instrument segmentation. Instrument category names are suboptimal as text prompts due to their coarse nature and lack of structural cues. Therefore, we introduce Collaborative Prompts that combine both category and part information of surgical instruments. To maximise the potential of these Collaborative Prompts and integrate instrument structure information with SAM's generic knowledge, we propose a part-to-whole collaborative prompting pipeline based on SAM, namely SP-SAM. Given a surgical image $I\in \mathbb{R}^{H\times W\times 3}$ of size $H\times W$ and Collaborative Prompts $T^{(c)}$ for an instrument category $c$, SP-SAM predicts the binary mask $M^{(c)}\in \{0,1\}^{H\times W}$ for the instrument.

With the Collaborative Prompts, instrument structure information can be easily integrated by establishing a category-part relation matrix $\mathcal{D}_{CP}\in \{0,1\}^{C\times P}$, where $C$ and $P$ denote the numbers of surgical instrument categories and instrument parts, respectively, and each element $d_{cp}$ in $\mathcal{D}_{CP}$ indicates the presence of part $p$ in category $c$. For instance, \textit{Monopolar Curved Scissors} (Fig.~\ref{fig:motivation}(e) middle instrument), composed of shaft and tip parts, would have 1s for these parts and 0s for absent parts like the wrist.  
SP-SAM leverages the expert knowledge on instrument structure $\mathcal{D}_{CP}$ for accurate surgical segmentation.

As shown in Fig.~\ref{fig:overall}, SP-SAM consists of four key components: 
(1) a frozen SAM Image Encoder that extracts image embeddings from the given image, 
(2) a Cross-Modal Prompt Encoder (Sec.~\ref{subsec:cmpe}) that extracts part embeddings from Collaborative Prompts and generates part sparse and dense embeddings through cross-modal interaction, (3) a Part-to-Whole Adaptive Fusion module (Sec.~\ref{sec:method-score}) that combines part sparse and dense embeddings into whole sparse and dense embeddings through Category Part Attention and Image Part Attention, considering category-specific and image-specific part contributions, respectively, and (4) a SAM Decoder for Hierarchical Decoding (Sec.~\ref{subsec:hd}) that decodes these embeddings into masks, thereby enhancing the model's comprehension of instruments both as a whole and at the part level.

\subsection{Cross-Modal Prompt Encoder}
\label{subsec:cmpe}

 \begin{figure*}[!t]
    \centering
      \includegraphics[width=1\textwidth]{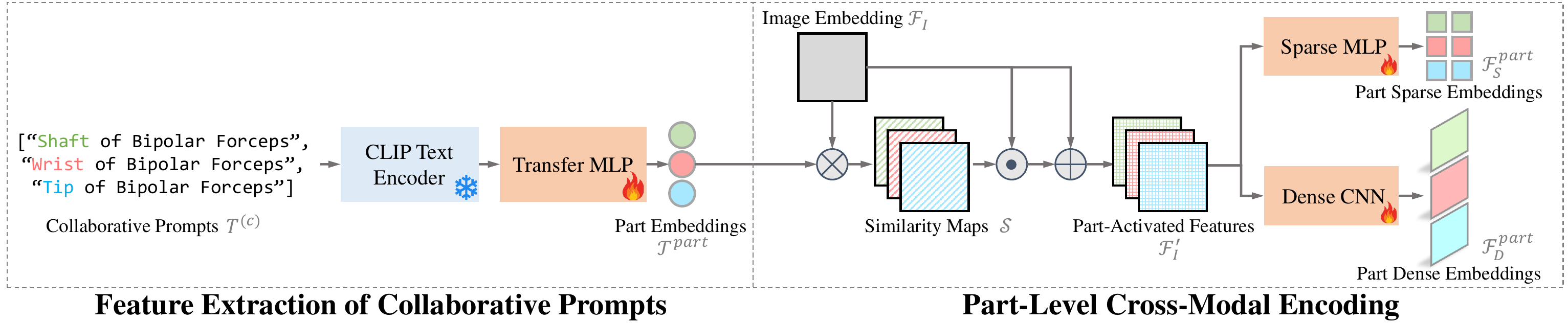}
    \caption{\textbf{Cross-Modal Prompt Encoder} consists of feature extraction of Collaborative Prompts and part-level cross-modal encoding.}
   \label{fig:crossmodal}
 \end{figure*}

The Cross-Modal Prompt Encoder takes the Collaborative Prompts and image embedding as input and performs cross-modal interaction between them via spatial attention, generating part sparse embeddings and part dense embeddings. As shown in Fig.~\ref{fig:crossmodal}, this process can be divided into two steps: feature extraction of Collaborative Prompts and part-level cross-modal encoding.

\textbf{Feature Extraction of Collaborative Prompts.}
We introduce a new type of text prompt for surgical instruments that collaboratively integrates both category and part information, namely Collaborative Prompts. Specifically, the Collaborative Prompts for an instrument of category $c$ are formulated into a set of texts containing all $P$ parts: $T^{(c)} = \{[part_p]\text{ of }[intrument\text{ }category_c]\}_{p=1}^P$, where $instrument\text{ }category_c$ and $part_p$ represent the names in text for instrument category $c$ and part $p$, respectively. 
Next, $T^{(c)}$ is encoded by the CLIP Text Encoder~\cite{clip} into text-based CLIP part embeddings $\mathcal{T}^{part}_{clip}\in \mathbb{R}^{P\times d_{clip}}$. 
A challenge here is the inherent distribution mismatch between the embedding spaces of SAM and CLIP. To transfer the CLIP text embeddings into SAM's embedding space, a tunable Transfer MLP is devised and applied to $\mathcal{T}^{part}_{clip}$, leading to the transferred embeddings for the parts, namely part embeddings $\mathcal{T}^{part}\in \mathbb{R}^{P\times d}$, where $d$ matches the number of embedding channels of SAM's image features.

\textbf{Part-Level Cross-Modal Encoding.}
In this step, the part embeddings $\mathcal{T}^{part}$ interact with the image embedding via spatial attention, and the obtained part-activated features are used to generate part sparse and dense embeddings. Specifically, the SAM Image Encoder extracts the image embedding $\mathcal{F}_I\in \mathbb{R}^{h\times w\times d}$, where $h\times w$ is the feature size. 
We then design a spatial attention mechanism by computing a similarity map for each part, leading to $\mathcal{S} = \mathcal{T}^{part} \times \mathcal{F}_I^\top \in \mathbb{R}^{P \times h\times w} $, where $\top$ denotes a transpose operator. These similarity maps serve as part-aware spatial attention to activate the image embedding, augmenting $\mathcal{F}_I$ into $\mathcal{F}'_{I} = \mathcal{S} \circ \mathcal{F}_I + \mathcal{F}_I \in \mathbb{R}^{P\times h\times w\times d}$, where $\mathcal{F}_I$ and $\mathcal{S}$ are broadcasted to the same size and $\circ$ denotes the Hadamard product. The part-activated features $\mathcal{F}'_I$, containing information of both the image and the Collaborative Prompts, are used to compute part sparse embeddings $\mathcal{F}_S^{part}\in \mathbb{R}^{P\times n\times d}$ and part dense embeddings $\mathcal{F}_D^{part}\in \mathbb{R}^{P\times h\times w\times d}$ with a two-layer MLP and a three-layer CNN, respectively. Here $n$ represents the number of sparse tokens for each part. These embeddings are then fed into the SAM Decoder to segment the corresponding instrument parts.

\subsection{Part-to-Whole Adaptive Fusion}
\label{sec:method-score}
\begin{figure}[!t]
\centering
\includegraphics[width=0.7\textwidth]{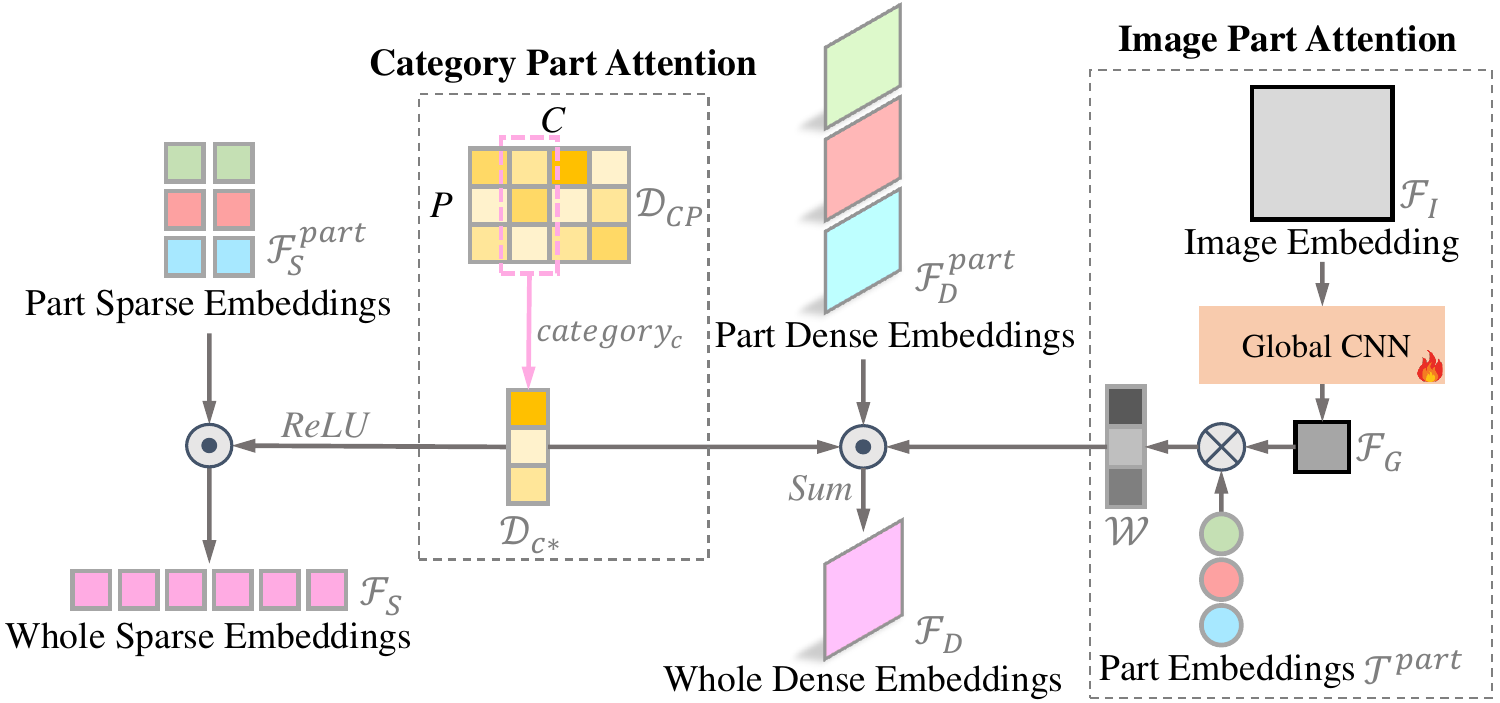}
\caption{\textbf{Part-to-Whole Adaptive Fusion Module} adaptively assembles the sparse and dense embeddings of the parts into the sparse and dense embeddings of the whole instrument via Category Part Attention and Image Part Attention.} 
\label{fig:p2w_fusion}
\end{figure}

In the Part-to-Whole Adaptive Fusion module, the sparse and dense embeddings for all parts are adaptively fused to form the whole sparse and dense embeddings, $\{\mathcal{F}_S, \mathcal{F}_D\}$, for the segmentation of the entire instrument. The adaptive fusion is achieved through \textbf{Category Part Attention} and \textbf{Image Part Attention}, as shown in Fig.~\ref{fig:p2w_fusion}. 
Specifically, the part sparse and dense embeddings consist of the prompt embeddings of all $P$ parts. However, as established in $\mathcal{D}_{CP}$, instruments of different categories encompass different part compositions. Therefore, we propose a Category Part Attention that utilises the part weights for the prompted category $c$ in $\mathcal{D}_{CP}$, \ie, $\mathcal{D}_{c*}=\{d_{cp}\}_{p=1}^P \in \mathbb{R}^{1\times P}$, as the weights to fuse the sparse and dense embeddings from the part level to the whole level. Note that  $\mathcal{D}_{CP}$ is initialised with 0s and 1s but is updated dynamically during model training.

While the Category Part Attention provides category-specific part weights, the presence and contribution of each part to an instrument can vary significantly across images due to different field-of-views and occlusion conditions. 
Therefore, it is necessary to adapt the part-to-whole fusion to the condition of each image. Accordingly, we propose Image Part Attention to compute image-specific part weights by learning a global descriptor of the image and computing its similarity with the part embeddings. 
Particularly, the global descriptor $\mathcal{F}_G\in \mathbb{R}^{1\times d}$ is learned from image embedding $\mathcal{F}_I$ with a Global CNN that consists of three convolutional layers and a linear layer. Then, image-specific part weights are computed as: $\mathcal{W}=\mathcal{F}_G\times \mathcal{T}^{part \top} \in \mathbb{R}^{1\times P}$.

Finally, given category-specific part weights $\mathcal{D}_{c*}$ and image-specific part weights $\mathcal{W}$, we fuse the sparse and dense embeddings $\{\mathcal{F}_S^{part}, \mathcal{F}_D^{part}\}$ of the parts into the sparse and dense embeddings $\{\mathcal{F}_S, \mathcal{F}_D\}$ of the whole instrument. Note that the matrices are all broadcasted to the same size prior to the Hadamard product. 
\begin{eqnarray}
        \mathcal{F}_S &=& \mathcal{F}^{part}_S \circ ReLU(\mathcal{D}_{c*}) \in \mathbb{R}^{P\times n \times d}, \\
    \mathcal{F}'_D &=& \mathcal{F}^{part}_D \circ \mathcal{D}_{c*} \circ  \mathcal{W} \in \mathbb{R}^{P \times h\times w\times d}, \\
    \mathcal{F}_D &=& \sum_{p=1}^P \mathcal{F}'_D\in \mathbb{R}^{h\times w\times d}.
\end{eqnarray}

\subsection{Hierarchical Decoding}
\label{subsec:hd}
The sparse and dense embeddings at both the whole level and the part level are fed into SAM Decoder for hierarchical decoding. This explicitly directs the model to learn both the overall structures as well as the part characteristics of surgical instruments. 
The final loss function thus comprises of the dice losses $\mathcal{L}_{D}$~\cite{vnet} for both the whole segmentation mask and the part segmentation masks:
\begin{eqnarray}
    &&\mathcal{L} = \mathcal{L}_{D}(M^{(c)}, G^{(c)}) + \sum_{p=1}^P d_{cp} \mathcal{L}_{D}(M^{(c)}_p, G^{(c)}_p), \qquad \label{loss} \\
    &&\mathcal{L}_{D}(M, G)=\frac{2\sum_{i}^{HW}m_ig_i}{\sum_i^{HW}m_i^2+\sum_i^{HW}g_i^2},
\end{eqnarray}
where $m_i$ and $g_i$ denote the predicted logit value and the ground-truth binary value at pixel $i$, respectively. $M^{(c)}$ and $\{M^{(c)}_p\}_{p=1}^P$ are the predicted masks of the instrument and its parts, respectively. $G^{(c)}$ and $\{G^{(c)}_p\}_{p=1}^P$ are the ground-truth masks of the instrument and its parts, respectively. 

\section{Experiments}
\label{sec:experiments}
\subsection{Datasets and Evaluation Metrics}
The effectiveness of SP-SAM is validated using the EndoVis2018~\cite{endovis2018} and EndoVis2017~\cite{endovis2017} datasets.
EndoVis2018 is composed of 11 training videos and four validation videos each with 149 frames, on which we follow the standard experiment and evaluation protocols defined in~\cite{ternausnet} and~\cite{isinet} to ensure a fair comparison with existing methods. EndoVis2017 contains eight training videos each with 255 frames and ten testing sequences with 900 frames in total. We adopt two evaluation protocols on EndoVis2017 for a fair comparison with different works: (1) average results of four-fold cross-validation, as per~\cite{ternausnet}; (2) training on the training set and reporting results on the test set, following the official code of~\cite{tpsis}. EndoVis2018 and EndoVis2017 offer annotations for five and four instrument parts, respectively, and both datasets include seven instrument categories. 

We adopt the standard evaluation metrics used in all existing works~\cite{isinet, baby,matis,surgicalsam,tpsis,mftapnet,dual_mf,trasetr,ternausnet}: Challenge IoU~\cite{endovis2017}, IoU~\cite{isinet}, and mean class IoU (mc IoU). 
Challenge IoU is computed only for the classes present in an image, whereas IoU considers all classes. 
We also report the IoU for each instrument category.

\subsection{Implementation Details}
Images from EndoVis2017 and EndoVis2018 are processed to a size of 1024$\times$1280, as per~\cite{ternausnet}. Data augmentation strategies are adopted following~\cite{ternausnet, matis}, which include random flipping, random scale and crop, random rotation, and colour jitter.
For Transfer MLP, Sparse MLP, Dense CNN, and Global CNN, their feature dimensions are set to 512, 256, 256, and 256, respectively. The number of sparse tokens per part $n$ is set to 2. In terms of training, we initialise SAM Image Encoder and SAM Decoder with SAM's pre-trained weights of the ViT-H version~\cite{vit}. We adopt CLIP Text Encoder of version ViT-L/14@336px, following~\cite{sam}. Our model keeps SAM Image Encoder, CLIP Text Encoder, and the output MLPs of SAM Decoder frozen, while updating the remaining weights using an Adam optimiser with a learning rate of 0.0001. To reduce computational load, we utilise pre-computed image embeddings, employing a batch size of 8. In practice, inspired by~\cite{surgicalsam}, we implement our model by inputting all categories into the model and differentiating the positive category (\ie, the prompted category) with negative categories via the positive and negative sparse embeddings of SAM. SP-SAM is trained and evaluated on an Nvidia Tesla V100 16GB GPU.

\subsection{Main Results}

\begin{figure*}[!t]
\centering
\includegraphics[width=0.98\textwidth]{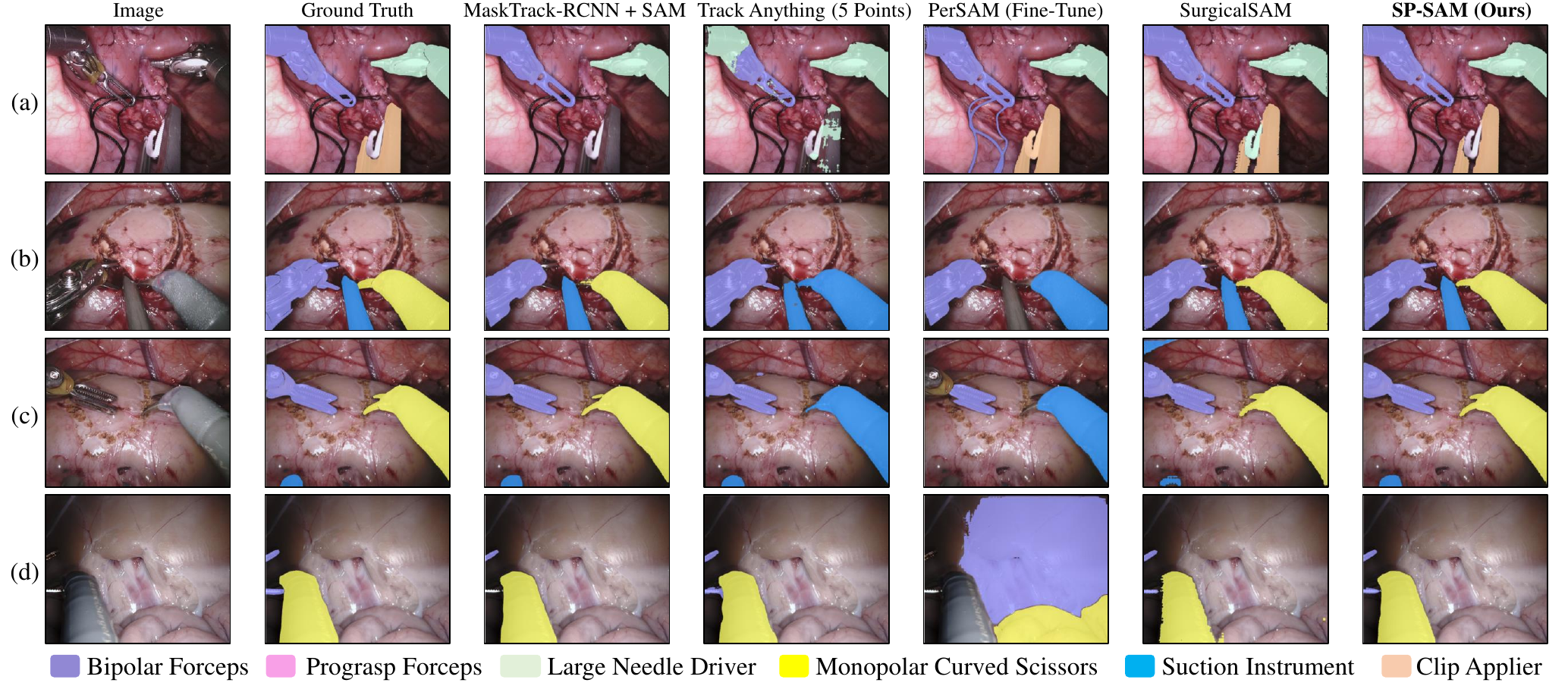}
\caption{Visual comparison of predicted masks by different methods.} 
\label{fig:vis_results}
\end{figure*}

We compare the performance of SP-SAM against existing methods on the EndoVis2018 and EndoVis2017 datasets, detailed in Table~\ref{tab:results_endovis2018} and Table~\ref{tab:results_endovis2017}, respectively. A visual comparison of the predictions is shown in Fig.~\ref{fig:vis_results} (More visualisations are provided in Supplemenary Materials.) The instrument categories include \textit{Bipolar Forceps (BF)}, \textit{Prograsp Forceps (PF)}, \textit{Large Needle Driver (LND)}, \textit{Suction Instrument (SI)}, \textit{Vessel Sealer (VS), Clip Applier (CA)}, \textit{Grasping Retractor (GR)}, \textit{Monopolar Curved Scissors (MCS)}, and \textit{Ultrasound Probe (UP)}. In our comparison, we divide existing methods into two categories: specialist models and SAM-based models. Notably, SP-SAM surpasses both existing fully-trained specialist models and efficient-tuning approaches based on SAM, yet at a substantially lower training cost in terms of tunable parameters. 

\begin{table*}[t]
\centering
\renewcommand{\arraystretch}{1}
\setlength{\tabcolsep}{3pt}
\begin{adjustbox}{width=\textwidth}
\begin{tabular}{l l c c c c c c c c c c c}
\hline
\rowcolor{black!10} & &  &  &  & \multicolumn{7}{c}{Instrument Categories} &  \\ \cline{6-12} 
\rowcolor{black!10} \multirow{-2}{*}{Method Category}  & \multirow{-2}{*}{Method}  & \multirow{-2}{*}{Challenge IoU} & \multirow{-2}{*}{IoU} & \multirow{-2}{*}{mc IoU} & BF & PF & LND & SI & CA & MCS & UP & \multirow{-2}{*}{\#T-Params} \\ \hline
\multirow{8}{*}{Specialist Model} & TernausNet~\cite{ternausnet} & 46.22 & 39.87 & 14.19 & 44.20 & 4.67 & 0.00 & 0.00 & 0.00 & 50.44 & 0.00 & 32.20M \\
  & MF-TAPNet~\cite{mftapnet}  & 67.87 & 39.14 & 24.68 & 69.23 & 6.10 & 11.68 & 14.00 & 0.91 & 70.24 & 0.57 & 37.73M \\
  & Dual-MF~\cite{dual_mf} & 70.40 & - & 35.09 & 74.10 & 6.80 & 46.00 & 30.10 & 7.60 & 80.90 & 0.10 & 203.80M \\
   & ISINet~\cite{isinet} &  73.03 & 70.94 & 40.21 & 73.83 & 48.61 & 30.98 & 37.68 & 0.00 & 88.16 & 2.16 & 162.52M\\
  & TraSeTr~\cite{trasetr} &  76.20 & -  & 47.71  & 76.30  & 53.30  & 46.50  & 40.60 & 13.90 & 86.20  & 17.15 & -  \\
    & S3Net~\cite{baby} & 75.81  &  74.02 &  42.58  & 77.22  &  50.87 &  19.83 &  50.59 &  0.00 &  92.12  & 7.44 & 68.41M  \\      
  & MATIS Frame~\cite{matis} & 82.37 & 77.01 & 48.65 & 83.35 & 38.82 & 40.19 & 64.49 & 4.32 & 93.18 & 16.17 & 68.72M\\ 
& TP-SIS~\cite{tpsis} & 84.92 & 83.61& 65.44 & 84.28 & 73.18 & 78.88 & 92.20 & 23.73 & 66.67 & 39.12 & 131.08M \\ 
  \hline
\multirow{10}{*}{SAM-based Model}  & MaskTrack-RCNN~\cite{masktrackrcnn} + SAM & 78.49 & 78.49 & 56.07 & 79.83 & 74.86 & 43.12 & \textbf{62.88} & 16.74  &  91.62 & 23.45 & 57.67M  \\
 & Mask2Former~\cite{m2f} + SAM & 78.72 & 78.72 & 52.50 & 85.95 & \textbf{82.31} & 44.08  & 0.00 & 49.80 & \textbf{92.17} & 13.18 & 68.72M \\
 & TrackAnything (1 Point)~\cite{trackanything}  &  40.36 &	38.38 & 20.62 & 30.20 & 12.87 & 24.46  & 9.17 & 0.19& 55.03 & 12.41 & -\\
 & TrackAnything (5 Points)~\cite{trackanything}  &  65.72 &  60.88 & 38.60 & 72.90 & 31.07 & \textbf{64.73}  & 10.24 & 12.28 & 61.05 & 17.93 & -\\
 & PerSAM (Zero-Shot)~\cite{persam} &  49.21 & 49.21 & 34.55 & 51.26 & 34.40 & 46.75 & 16.45 & 15.07 & 52.28 & 25.62 & - \\
 & PerSAM (Fine-Tune)~\cite{persam} &  52.21 & 52.21 & 37.24 & 57.19 & 36.13 & 53.86 & 14.34 & 25.94 & 54.66 & 18.57 & 2 \\
 & SurgicalSAM~\cite{surgicalsam} & 80.33 & 80.33 & 58.87 & 83.66 & 65.63 & 58.75  & 54.48 & 39.78 & 88.56 & 21.23 & 4.65M\\ 
  & \textbf{SP-SAM (Ours)} & \textbf{84.24} & \textbf{84.24} & \textbf{65.71} & \textbf{87.60} & 65.07 & 61.95 & 58.30 & \textbf{59.96}  & 92.08 & \textbf{34.99} & 8.62M \\ 
  \cline{2-13} 
 & GT Centroid + SAM & 60.26 & 60.26 & 63.34 & 44.35 & 65.92 & 30.99 & 87.14 & 69.69 & 80.04 & 65.26  & -\\
 & GT Bbox + SAM & 88.04 & 88.04 & 84.23 & 87.10 & 86.81 & 72.23 & 91.21 & 75.91 & 93.08 & 83.24 & -\\
\hline
\end{tabular}
\end{adjustbox}
\caption{Comparison of results on the EndoVis2018 dataset. \#T-Params denotes the number of tunable parameters.}
\label{tab:results_endovis2018}
\end{table*}

\begin{table*}[t]
\centering
\renewcommand{\arraystretch}{1}
\setlength{\tabcolsep}{4.5pt}
\begin{adjustbox}{width=\textwidth}
\begin{tabular}{l l c c c c c c c c c c}
\hline
 \rowcolor{black!10} &  &  &  &  & \multicolumn{7}{c}{Instrument Categories} \\ \cline{6-12} 
\rowcolor{black!10} \multirow{-2}{*}{Method Category} & \multirow{-2}{*}{Method} & \multirow{-2}{*}{Challenge IoU} & \multirow{-2}{*}{IoU} & \multirow{-2}{*}{mc IoU} & BF & PF & LND & VS & GR & MCS & UP \\ \hline
\rowcolor{orange!10} & & \multicolumn{10}{c}{\textbf{Cross-Fold Average Results}} \\
 \hline
\multirow{8}{*}{Specialist Model} & TernausNet~\cite{ternausnet} & 35.27 & 12.67 &  10.17 & 13.45 & 12.39 & 20.51 & 5.97 & 1.08 & 1.00 & 16.76 \\
  & MF-TAPNet~\cite{mftapnet} & 37.25 & 13.49 & 10.77 & 16.39 &  14.11 & 19.01&  8.11 &  0.31   & 4.09  & 13.40 \\
   & Dual-MF~\cite{dual_mf} & 45.80 & - & 26.40 & 34.40 & 21.50 & 64.30 & 24.10 & 0.80 & 17.90 & 21.80 \\
  & ISINet~\cite{isinet} & 55.62 & 52.20 & 28.96 & 38.70 & 38.50 & 50.09 & 27.43 & 2.10 & 28.72 & 12.56 \\ 
  & TraSeTr~\cite{trasetr} & 60.40 & - & 32.56 & 45.20  &   56.70  &  55.80 &  38.90  &  11.40 &  31.30 &  18.20  \\
    & S3Net~\cite{baby} & 72.54 &  71.99 & 46.55 & 75.08
 & 54.32  & 61.84 & 35.50 &  27.47  & 43.23  & 28.38 \\      
  & MATIS Frame~\cite{matis} & 68.79 & 62.74 & 37.30 & 66.18 & 50.99 & 52.23 & 32.84 & 15.71 & 19.27 & 23.90 \\
& TP-SIS~\cite{tpsis}  & 63.37  & 63.37  & 52.74 & 66.42 & 45.46 & 75.20 & 73.44 & 29.95 & 44.02 & 34.67 \\ 
  \hline
\multirow{9}{*}{SAM-based Model}  &  Mask2Former~\cite{m2f} + SAM & 66.21 & 66.21 & 55.26 & 66.84 & \textbf{55.36} & 83.29 & 73.52 & 26.24 & 36.26 & 45.34 \\
 & TrackAnything (1 Point)~\cite{trackanything} & 54.90 & 52.46 & 55.35 & 47.59 & 28.71 & 43.27 & 82.75 & 63.10 & 66.46 & 55.54  \\
& TrackAnything (5 Points)~\cite{trackanything} & 67.41 & 64.50 & 62.97 & 55.42 & 44.46 & 62.43 & \textbf{83.68} & 62.59 & 67.03 & \textbf{65.17} \\
 & PerSAM (Zero-Shot)~\cite{persam} & 42.47 & 42.47 & 41.80 & 53.99 & 25.89 & 50.17 & 52.87 & 24.24 & 47.33 & 38.16 \\
& PerSAM (Fine-Tune)~\cite{persam}&  41.90 & 41.90 & 39.78 & 46.21 & 28.22 & 53.12 & 57.98 & 12.76 & 41.19 & 38.99\\
 & SurgicalSAM~\cite{surgicalsam} & 69.94 & 69.94 & 67.03 & 68.30 & 51.77 & 75.52 & 68.24 & 57.63 & \textbf{86.95 }& 60.80 \\ 
 & \textbf{SP-SAM (Ours)}& \textbf{73.94} & \textbf{73.94} & \textbf{71.06} & \textbf{68.89} & 53.16 & \textbf{83.80} & 73.20 & \textbf{72.40} & 84.91 & 61.05\\
 \cline{2-12}
 & GT Centroid + SAM & 44.42 & 44.42 & 54.41 & 63.42 & 36.03 & 22.57 & 54.21 & 75.18 & 70.17 & 59.25 \\
 & GT Bbox + SAM & 76.31 & 76.31 & 81.18 & 89.36 & 73.44 & 67.67 & 90.04 & 87.79 & 94.03 & 65.91 \\
\hline
\rowcolor{orange!10} & & \multicolumn{10}{c}{\textbf{Test Set Results}} \\
\hline
Specialist Model & TP-SIS~\cite{tpsis} & 79.90 & 77.83 & \textbf{56.22} & 68.58 & 73.52 & \textbf{92.74} & \textbf{83.90} & 0.13 & \textbf{74.70} & 0.00  \\ 
SAM-based Model & \textbf{SP-SAM (Ours)} & \textbf{82.01} & \textbf{82.01} & 56.00 & \textbf{81.64} & \textbf{74.06} & 91.42 & 72.00 & \textbf{0.84} & 72.06 & 0.00 \\
\hline
\end{tabular}
\end{adjustbox}
\caption{Comparison of results on the EndoVis2017 Dataset.}
\label{tab:results_endovis2017}
\end{table*}

In terms of  zero-shot performance of SAM, the methods using bounding boxes as prompts (MaskTrack-RCNN~\cite{masktrackrcnn}/Mask2Former~\cite{m2f} + SAM) in general outperform those using point prompts (TrackAnything~\cite{trackanything}) and image prompts (PerSAM~\cite{persam}). However, their performances are still inferior to tuning-based methods. Additionally, they rely on a well-trained detector for bounding box prediction, resulting in a considerable increase in the number of training parameters and higher pipeline complexity. In contrast, tuning-based methods for SAM, including both SurgicalSAM~\cite{surgicalsam} and SP-SAM, adopt an end-to-end efficient tuning pipeline with minimal training parameters, boosting both training efficiency and segmentation performance.

Our method demonstrates considerable enhancement over the existing efficient-tuning approach, SurgicalSAM~\cite{surgicalsam}, on both datasets.
It shows an improvement of 3.91 and 4.00 in terms of Challenge IoU on EndoVis2018 and EndoVis2017, respectively. Unlike SurgicalSAM, which relies on category ID prompting, our approach is prompted by text, leveraging the extensive information in natural language expressions and pre-trained language models. Additionally, our method significantly outperforms SurgicalSAM in mean class IoU (mc IoU), by a gain of 6.84 for EndoVis2018 and 4.03 for EndoVis2017, indicating superior discrimination of instruments across different categories. This enhancement is largely attributed to our part-to-whole collaborative prompting mechanism that explicitly directs the model to identify the internal structures of instruments and concentrate on the part-level details, in contrast to SurgicalSAM which treats each instrument as a single entity.

In addition, we compare SP-SAM with two oracle scenarios that employ ground-truth centroids and bounding boxes as prompts for SAM. Remarkably, SP-SAM achieves significantly better results than those obtained with ground-truth centroids. Moreover, SP-SAM's performance closely approaches that of the oracle setting with ground-truth bounding boxes, with only a gap of 3.80 and 2.37 in terms of Challenge IoU for EndoVis2018 and EndoVis2017, respectively, yet our method requires significantly less prompting effort without the need for manual per-frame bounding box guidance.

Finally, SP-SAM achieves superior or competitive performance compared to SOTA specialist models while using significantly fewer tunable parameters. On EndoVis2018, our method surpasses TP-SIS~\cite{tpsis} in both IoU and mc IoU, despite TP-SIS utilising 15 times more tunable parameters than ours (131.08M for TP-SIS vs. 8.62M for SP-SAM). On EndoVis2017, for cross-fold averages, we compare with the reproduced results of TP-SIS using the official code; for test set results, we compare with the results reported in \cite{tpsis}. Our methods exhibit improvements in both settings, with a more substantial enhancement in cross-fold averages, a more robust evaluation protocol, further affirming our superiority. In contrast to TP-SIS which utilises instrument part masks straightforwardly as training supervisory signals, our method more effectively integrates instrument structure knowledge and explicitly addresses category-part and image-part relationships. This allows for a more accurate comprehension of instruments, encompassing both their structures and finer details. Similarly, the comparison between SP-SAM and its variant without Part-to-Whole Adaptive Fusion (model C in Table~\ref{tab:ablat}) in our ablation study also highlights the improvement our method offers over straightforward use of instrument part masks.

Figure \ref{fig:vis_results} showcases a visual comparison of predicted masks by different methods, highlighting that SP-SAM clearly outperforms existing SAM-based methods in segmentation quality. Notably, SP-SAM outperforms SurgicalSAM, which tends to misidentify instrument categories (Fig.~\ref{fig:vis_results}(a)), predict incomplete masks with missed critical parts like instrument tips (Fig.~\ref{fig:vis_results}(b)), and generate rugged edges (Fig.~\ref{fig:vis_results}(c) and (d)). Owing to the part-to-whole collaborative prompting mechanism, SP-SAM precisely captures fine-grained details such as edges and challenging areas. This precision, especially in identifying edges and tips, is crucial for ensuring safety in surgical settings.

\subsection{Ablation Study}

\textbf{Ablation Study of Key Components.} We conduct an ablation study on both EndoVis2018 and EndoVis2017 to investigate the effect of the proposed Collaborative Prompts, Part-to-Whole Adaptive Fusion, and Hierarchical Decoding. The results are reported in Table~\ref{tab:ablat}. 
\textbf{Model A}, the baseline, utilises text prompts of category names, which are encoded by the Cross-Model Prompt Encoder into sparse and dense embeddings for the SAM Decoder to generate instrument masks. Model A is then progressively augmented with our proposed modules. In \textbf{model B}, Collaborative Prompts are encoded by the Cross-Modal Prompt Encoder into sparse and dense embeddings for individual parts. These embeddings of different parts are then combined into whole sparse and dense embeddings in a straightforward manner, where the sparse embeddings of all parts are concatenated and the dense embeddings of all parts are summed. Subsequently, these whole sparse and dense embeddings are decoded by the SAM Decoder into instrument masks. Building upon model B, \textbf{model C} additionally incorporates Hierarchical Decoding, enabling the decoding of both whole and part sparse and dense embeddings into their respective masks. Following model C, we investigate the impact of Category Part Attention and Image Part Attention in the Part-to-Whole Adaptive Fusion module. In \textbf{model D}, we set image-specific part weights $\mathcal{W}$ to an all-one vector, while in \textbf{model E}, we fix $\mathcal{D}_{CP}$ as an all-one matrix. Finally, \textbf{model F} represents our proposed SP-SAM.

\begin{table}[t]
\renewcommand{\arraystretch}{1}
\setlength{\tabcolsep}{3pt}
\centering
\begin{adjustbox}{width=0.9\textwidth}
\begin{tabular}{ c | c  c  c c  |c c | c c }
\hline
\rowcolor{black!10} &  & \multicolumn{2}{c}{Part-to-Whole Fusion} &  & \multicolumn{2}{c|}{EndoVis2018} &  \multicolumn{2}{c}{EndoVis2017} \\ \cline{3-4} \cline{6-7} \cline{8-9}
\rowcolor{black!10}  \multirow{-2}{*}{Model}   &  \multirow{-2}{*}{Collab. Prompts}  & Category Att.& Image Att.& \multirow{-2}{*}{Hier. Decod.} & Challenge IoU & mc IoU  & Challenge IoU & mc IoU  \\ \hline
A &  &    &&   &  81.08 & 61.14 & 67.69 & 63.59 \\ 
 \hline
B & \ding{51}   &    &&  & 81.90 &  62.23 & 68.97 & 65.30\\ 
C & \ding{51}   &    && \ding{51}  &  82.36 & 61.26 & 71.64 & 66.21\\ 
D & \ding{51}  &  \ding{51} && \ding{51}  &   82.64 & 63.24  & 72.55 & 67.87\\
E & \ding{51}  &  &\ding{51} & \ding{51}  &  82.98  &  65.37 & 72.48 & 68.21\\
F & \ding{51}  &  \ding{51} &\ding{51} & \ding{51}  &   \textbf{84.24} & \textbf{65.71} & \textbf{73.94} & \textbf{71.06}\\
\hline
\end{tabular}
\end{adjustbox}
\caption{Ablation study of the key components of SP-SAM.}
\label{tab:ablat}
\end{table}

\begin{figure*}[t]
\centering
\includegraphics[width=0.98\textwidth]{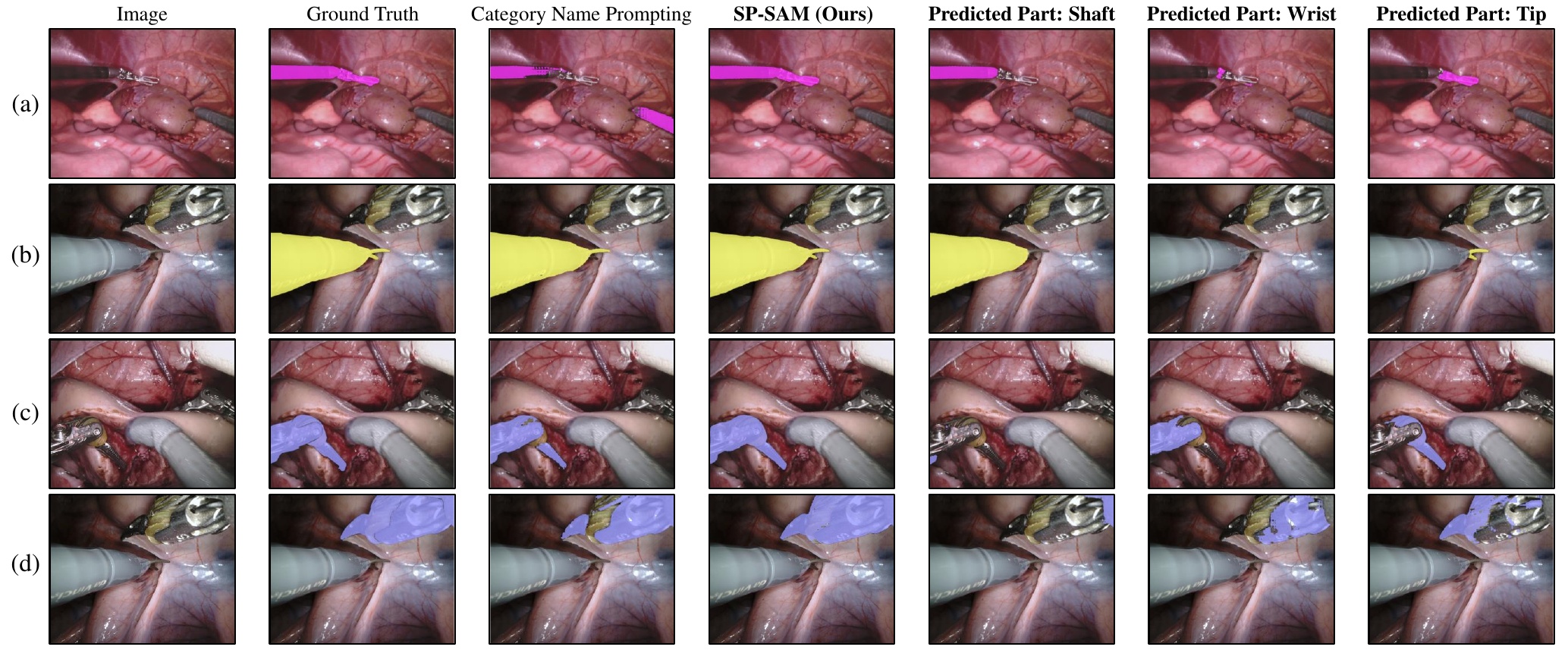}
\caption{Visual comparison: SP-SAM vs. instrument category name prompting.} 
\label{fig:vis_abl}
\end{figure*}

In general, it can be seen that each module individually enhances the Challenge IoU and mc IoU scores with the most significant improvements observed when all modules are integrated. 
In particular, in EndoVis2018, after incorporating Collaborative Prompts (model B), adding Hierarchical Decoding (model C) shows a marginal improvement in Challenge IoU and a decrease in mc IoU. This confirms that employing surgical part labels as auxiliary signals in a straightforward multi-task learning manner does not yield substantial enhancements. The real breakthrough comes with the proposed Part-to-Whole Adaptive Fusion in conjunction with Hierarchical Decoding (models D, E, and F), unlocking the full potential. A visual comparison of SP-SAM with the baseline of category name prompting is presented in Fig.~\ref{fig:vis_abl}. Prompting without part information results in a substantial loss of details in the predicted masks. In contrast, SP-SAM excels in recognising all instrument parts, particularly with intricate elements like tips (Fig.~\ref{fig:vis_abl}(a) and (b)) and areas with varying materials (Fig.~\ref{fig:vis_abl}(c) and (d)).

\textbf{Ablation Study of Sparse and Dense Embeddings}. We conduct an ablation study on sparse and dense embeddings on both EndoVis2018 and EndoVis2017 datasets, as shown in Table \ref{tab:ablation_prompt}. The study evaluates the model's performance when using only sparse or only dense embeddings. In the sparse-only model, dense embeddings are replaced with the no-mask embeddings from SAM's pre-trained weights, and in the dense-only model, sparse embeddings are set to be empty tensors. Table \ref{tab:ablation_prompt} reveals that removing either dense or sparse embeddings decreases performance, with the removal of dense embeddings causing a more significant decline. This suggests that dense embeddings are more crucial than sparse embeddings. This aligns with our expectations, since in SAM dense embeddings function as masks, holding more information than the point-based sparse embeddings, thus guiding more accurate decoding. Our method optimally leverages both sparse and dense embeddings and achieves the best results.

\begin{table}[t]
\renewcommand{\arraystretch}{1}
\setlength{\tabcolsep}{10pt}
\centering
\begin{adjustbox}{width=0.8\textwidth}
\begin{tabular}{l | c c | c c}
\hline
\rowcolor{black!10} & \multicolumn{2}{c|}{EndoVis2018} &  \multicolumn{2}{c}{EndoVis2017} \\
 \cline{2-5}
\rowcolor{black!10} \multirow{-2}{*}{Method}  & Challenge IoU & mc IoU & Challenge IoU & mc IoU \\ 
\hline
Sparse-Only  & 76.71 & 57.17 & 67.15 & 64.18 \\
Dense-Only & 83.09 & 63.81 & 72.43 & 66.61 \\
\textbf{SP-SAM (Ours)} & \textbf{84.24} & \textbf{65.71}  & \textbf{73.94} & \textbf{71.06}   \\
\hline
\end{tabular}
\end{adjustbox}
\caption{Ablation study of sparse and dense embeddings.}
\label{tab:ablation_prompt}
\end{table}

\textbf{Impact of Number of Tokens.} Details are in Supplementary Materials.

\section{Conclusion}
\label{sec:conclusion}

In this paper, we present SP-SAM, an efficient-tuning approach of SAM for text promptable surgical instrument segmentation. It leverages a part-to-whole collaborative prompting mechanism to address the challenge of complex structures and fine-grained details in surgical instruments. Specifically, Collaborative Prompts are devised to describe surgical instruments at both category and part levels. Moreover, the proposed Cross-Modal Prompt Encoder, Part-to-Whole Adaptive Fusion, and Hierarchical Decoding modules learn discriminative representations of instrument parts and adaptively assemble them for accurate instrument segmentation. Experiments on both EndoVis2018 and EndoVis2017 datasets demonstrate that SP-SAM outperforms both specialist methods and SAM-based methods while only tuning a small number of parameters. Our method demonstrates the great potential of efficiently adapting foundation models for highly specialised tasks and offers valuable insights into the segmentation of challenging targets. In the future, our method can be further improved by exploring other forms of text prompts, incorporating temporal cues, and tackling background targets such as human tissues.

\bibliographystyle{splncs04}
\bibliography{main}

\end{document}